\documentclass[letterpaper, 10 pt, conference]{ieeeconf}  

\IEEEoverridecommandlockouts                              

\usepackage{cite}
\usepackage{amsmath,amssymb,amsfonts}
\usepackage{algorithmic}
\usepackage{graphicx}
\usepackage{textcomp}
\usepackage[dvipsnames,x11names]{xcolor}

\usepackage{times}
\usepackage{epsfig}
\usepackage{graphicx}
\usepackage{amsmath}
\usepackage{amssymb}
\usepackage{textcomp}
\usepackage{multirow}
\usepackage{hhline}
\usepackage{subcaption}
\usepackage{verbatim}
\usepackage{tabularx}
\usepackage[group-separator={,},mode=text]{siunitx}

\let\labelindent\relax
\usepackage{enumitem}

\usepackage{gensymb}
\usepackage{flushend}
\usepackage{xcolor}

\usepackage[ruled,vlined]{algorithm2e}

\usepackage[labelformat=simple]{subcaption}
\DeclareCaptionLabelSeparator{periodspace}{.\quad}
\definecolor{green1}{RGB}{14, 81, 7}
\definecolor{green2}{RGB}{63, 125, 49}
\definecolor{green3}{RGB}{123, 175, 112}

\renewcommand{\thetable}{\arabic{table}}
\usepackage{booktabs}
\usepackage{array}

\def\BibTeX{{\rm B\kern-.05em{\sc i\kern-.025em b}\kern-.08em
    T\kern-.1667em\lower.7ex\hbox{E}\kern-.125emX\}}}

\makeatletter
\let\NAT@parse\undefined
\makeatother
\usepackage{hyperref}
\hypersetup{
  colorlinks=true,
  citecolor=SpringGreen4
}
\usepackage{cleveref}

\title{\LARGE \bf
Woodscape Fisheye Object Detection for Autonomous Driving \\ -- CVPR 2022 OmniCV Workshop Challenge
}

\author{Saravanabalagi Ramachandran$^{1}$, Ganesh Sistu$^{2}$, Varun Ravi Kumar$^{3}$, John McDonald$^{1}$ and Senthil Yogamani$^{3}$
\thanks{$^{1}$Saravanabalagi Ramachandran and John McDonald are with Lero - the Irish Software Research Centre and the Department of Computer Science, Maynooth University, Maynooth, Ireland {\tt \{saravanabalagi.ramachandran, john.mcdonald\}@mu.ie}.}
\thanks{$^{2}$Ganesh Sistu is with Valeo Vision Sytems, Ireland. {\tt \{ganesh.sistu \}@valeo.com}.}
\thanks{$^{3}$Varun Ravi Kumar and Senthil Yogamani are with Qualcomm. {\tt \{vravikum, syogaman\}@qualcomm.com}.}
}
\begin{document}
\maketitle
\thispagestyle{empty}
\pagestyle{empty}
\bstctlcite{IEEEexample:BSTcontrol}
\begin{abstract}

Object detection is a comprehensively studied problem in autonomous driving. However, it has been relatively less explored in the case of fisheye cameras. The strong radial distortion breaks the translation invariance inductive bias of Convolutional Neural Networks. Thus, we present
the WoodScape fisheye object detection challenge for autonomous driving which was held as part of the CVPR 2022 Workshop on Omnidirectional Computer Vision (OmniCV). This is one of the first competitions focused on fisheye camera object detection.
We encouraged the participants to design models which work natively on fisheye images without rectification. We used CodaLab to host the competition based on the publicly available WoodScape fisheye dataset. 
In this paper, we provide a detailed analysis on the competition
which attracted the participation of 120 global teams and a total of 1492 submissions. We briefly discuss the details of the winning methods and analyze their qualitative and quantitative results. 
\end{abstract}

\section{Introduction}
\begin{figure}[tb]
\centering
\includegraphics[trim={ 0 0 2cm 0},clip, width=\columnwidth]{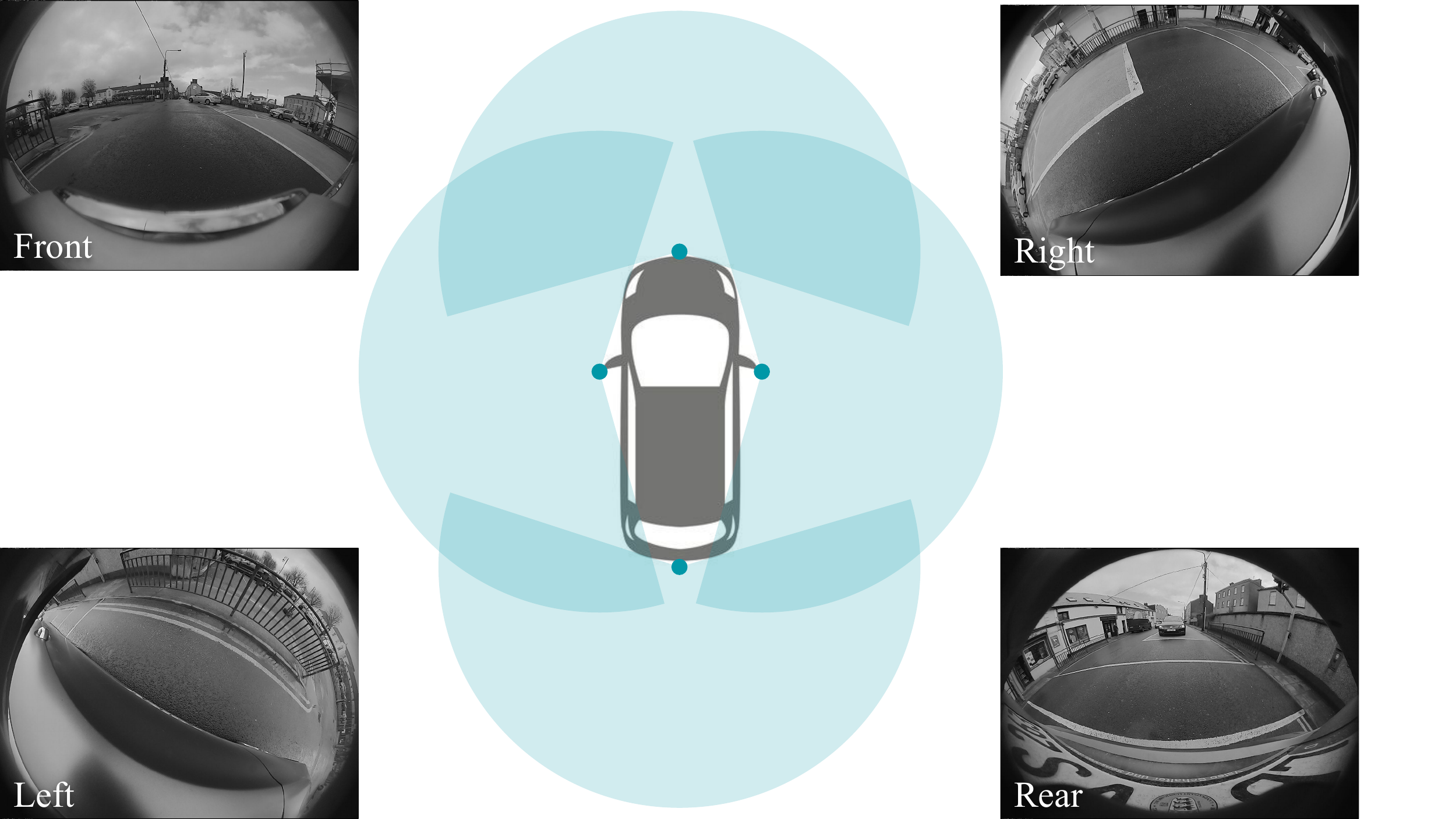}
\caption{Illustration of a typical automotive surround-view system consisting of four fisheye cameras located at the front, rear, and on each wing mirror covering the entire $360^\circ$ around the vehicle.} 
\label{fig:svs}
\end{figure}

Autonomous Driving is a challenging problem and it requires multiple sensors handling different aspects and robust sensor fusion algorithms which combine the senor information effectively \cite{yadav2020cnn, mohapatra2021bevdetnet, dasgupta2022spatio}. 
Surround-view systems employ four sensors to create a network with large overlapping zones to cover the car's near-field area \cite{eising2021near, kumar2022surround}. For near-field sensing, wide-angle images reaching $180^\degree$ are utilized. Any perception algorithm must consider the substantial fisheye distortion that such camera systems produce. Because most computer vision research relies on narrow field-of-view cameras with modest radial distortion, this is a substantial challenge. However, because camera systems are now more commonly used, development in this field has been attained. \autoref{fig:svs} illustrates the typical automotive surround-view camera system comprising of four fisheye cameras covering the entire $360^\circ$ around the vehicle. 
Most commercial cars have fisheye cameras as a primary sensor for automated parking. Rear-view fisheye cameras have become a typical addition in low-cost vehicles for dashboard viewing and reverse parking. Despite its abundance, there are just a few public databases for fisheye images, so relatively little research is conducted. One such dataset is the Oxford RobotCar \cite{maddern20171} a large-scale dataset focusing on the long-term autonomy of autonomous vehicles. The key responsibilities of this dataset, which enables research into continuous learning for autonomous cars and mobile robotics, are localization and mapping. It includes approximately 100 repetitions of a continuous route around Oxford, UK, collected over a year and commonly used for long-term localization and mapping.\par

\begin{figure*}[tb]
\centering
\includegraphics[width=\textwidth]{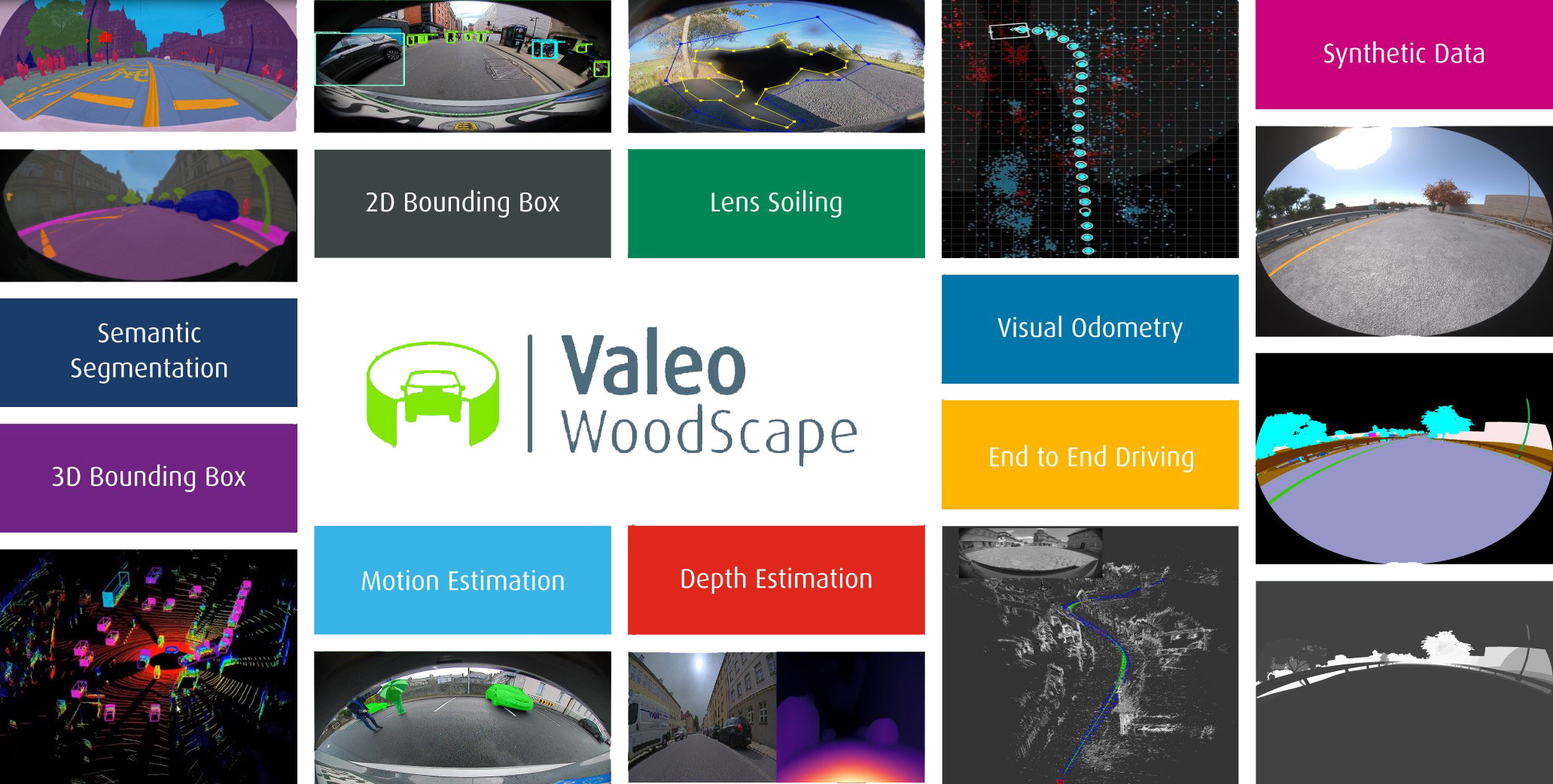}
\caption{Illustration of various perception tasks in WoodScape dataset.} 
\label{fig:woodscape-tasks}
\end{figure*}

WoodScape \cite{yogamani2019_woodscape} is a large dataset for $360^\degree$ sensing around an ego vehicle with four fisheye cameras. It is designed to complement existing automobile datasets with limited FOV images and encourage further research in multi-task multi-camera computer vision algorithms for self-driving vehicles. It is built based on industrialization needs addressing the diversity challenges \cite{uricar2019challenges}.  The dataset sensor configuration consists of four surround-view fisheye cameras sampled randomly. The dataset comprises labels for geometry and segmentation tasks, including semantic segmentation, distance estimation, generalized bounding boxes, motion segmentation, and a novel lens soiling detection task (shown in \autoref{fig:woodscape-tasks}).

Instead of naive rectification, the WoodScape pushes researchers to create solutions that can work directly on raw fisheye images, modeling the underlying distortion. WoodScape dataset (public and private versions) has enabled research in various perception areas such as object detection~\cite{dahal2021online, rashedfisheyeyolo, rashed2021generalized,  yahiaoui2019optimization}, trailer detection \cite{dahal2019deeptrailerassist}, soiling detection~\cite{uricar2021let, das2020tiledsoilingnet, uricar2019desoiling}, semantic segmentation \cite{cheke2022fisheyepixpro, sobh2021adversarial, dahal2021roadedgenet, klingner2022detecting, rashed2019motion}, weather classification \cite{dhananjaya2021weather}, depth prediction \cite{kumar2018monocular, kumar2018near, kumar2021svdistnet, kumar2020unrectdepthnet, kumar2020fisheyedistancenet, kumar2021fisheyedistancenet++, kumar2020syndistnet}, moving object detection \cite{siam2018modnet, yahiaoui2019fisheyemodnet, rashed2019motion, mohamed2021monocular}, SLAM \cite{tripathi2020trained, gallagher2021hybrid} and multi-task learning \cite{leang2020dynamic, kumar2021omnidet}.
SynwoodScape \cite{sekkat2022synwoodscape} is a synthetic version of the Woodscape dataset.

This paper discusses the second edition of WoodScape dataset challenge focused on the object detection task. The results of the first edition of the WoodScape challenge are discussed in \cite{ramachandran2021woodscape}. It is organized as part of the CVPR 2022 OmniCV workshop [\href{https://sites.google.com/view/omnicv2022/program}{link}]. Section \ref{sec:challenge} discusses the challenge setup including metrics and conditions. Section \ref{sec:outcome} discusses the participation information and the details of the winning solutions. Finally, Section \ref{sec:conc} provides concluding remarks. \par

\section{Challenge} \label{sec:challenge}

\begin{figure*}[tb]
\centering
\includegraphics[width=\textwidth]{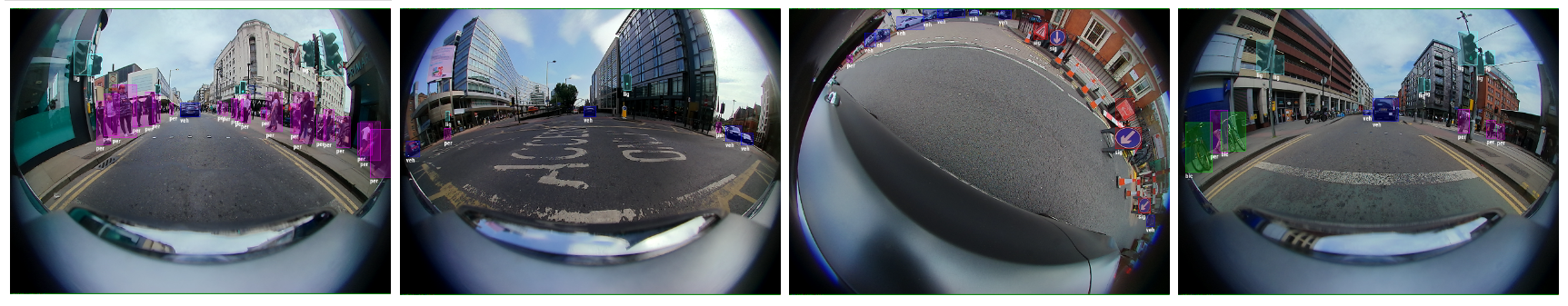}
\caption{Illustration of object detection annotations.}
\label{fig:segmentation-task}

\end{figure*}

The intent of the competition is to evaluate fisheye object detection techniques encouraging novel architectures for processing fisheye images, adaptations to existing network architectures, and handling of radial distortion without rectification.
To illustrate the difficulty of object detection in fisheye images, we show sample images and their ground truth in \autoref{fig:segmentation-task}. It is easy to notice that the radial distortion is much higher than other public datasets.  Further, fisheye images have overwhelming characteristics
including huge scale variance, complicated background filled with distractors, non-linear distortions, which pose enormous challenges for general object detectors based on common Convolutional Neural Networks (CNN) architectures.

Train and test split of the dataset is illustrated in \autoref{tab:train_test_split}. Class labels and their corresponding objects are listed in  \autoref{tab:label_mapping}. We choose the five important classes based on automated driving importance and the class frequency. Thus the challenge has 5 important classes instead of the 40 overall annotated classes. 
To prevent participants from overfitting their model on the test data using information available in the leaderboard, the challenge was held in 2 phases: Dev Phase and Test Phase. Details of start and end dates and the duration is shown in table \autoref{tab:challenge_phases}. During the Dev Phase, evaluation was done on a subset of the original test set to prevent participants from overfitting their model on the test data. The subset of random 1000 images is the same for all participants, however, the subset was changed every week. During the Test Phase, the submissions are evaluated on the entire test set.

\begin{table}
\centering
\begin{tabular}{cccc} \toprule
Phase & Start Date & End Date & Duration \\\midrule
Dev & April 15 & June 3 & 50 days \\
Test & June 4 & June 5 & 2 days \\\bottomrule
\end{tabular}
\caption{Details of start and end dates of the two phases of the challenge.}
\label{tab:challenge_phases}
\end{table}

\begin{table}
\centering
\begin{tabular}{ccc} \toprule
Split & Images & Percent \\\midrule
Training Set & 8234 & 82.34\% \\
Test Set & 1766 & 17.66\% \\\midrule
Total & 10000 & 100.00\% \\\bottomrule
\end{tabular}
\caption{Train and test data split of the dataset samples.}
\label{tab:train_test_split}
\end{table}

\begin{table}
\centering
\begin{tabular}{ccc} \toprule
Label & Description \\\midrule
0 & Vehicles \\
1 & Person \\
2 & Bicycle \\
3 & Traffic Light \\
4 & Traffic Sign \\\bottomrule
\end{tabular}
\caption{Class labels 0-4 and their corresponding objects.}
\label{tab:label_mapping}
\end{table}

\subsection{Metrics}

Mean Average Precision (mAP) is a standard evaluation metric for object detection tasks, which first computes the Average Precision (AP) for each class and then computes the average over classes. 

For each image in the test set, the correspondence for a ground truth bounding box is established by choosing the bounding box that has the maximum IoU among all propose bounding boxes. Each bounding box is then categorized as either TP (True Positive) or FP (False Positive). Correspondences matching is done without replacement to avoid one-to-many correspondences. A bounding box proposed is considered as an TP if it has an IoU (Intersection over Union) of more than the IoU Threshold 0.5 with the corresponding ground truth bounding box, and is marked an FP if less. Intersection of Union for two bounding boxes is defined as:

\begin{equation}
\text{IoU} = \frac{\text{Area of Intersection}}{\text{Area of Union}}
\end{equation}

The competition entries in CodaLab were evaluated and ordered based on mAP, averaged to all 5 classes. Along with the overall rank based on mAP, rank based on AP for individual classes are also displayed in the leaderboard.

\subsection{Reward}
The winning team will receive €1,000 through sponsorship from \href{https://www.lero.ie}{Lero} and will be offered to present in-person or virtually in the OmniCV 3rd Workshop held in conjunction with IEEE Computer Vision and Pattern Recognition (CVPR) 2022.

\subsection{Conditions}

Competition rules allowed the teams to make use of any other public datasets for pre-training. There were no restrictions on computational complexity as well. There is no limit on team size but there is a limit of 10 submissions per day and 150 submissions in total for a team.
For the Test phase, submissions were limited to a maximum of 2. 
Valeo employees or their collaborators who have access to the full WoodScape dataset were not allowed to take part in this challenge.

\section{Outcome} \label{sec:outcome}

\begin{figure*}[tb]
\centering
\includegraphics[width=\textwidth]{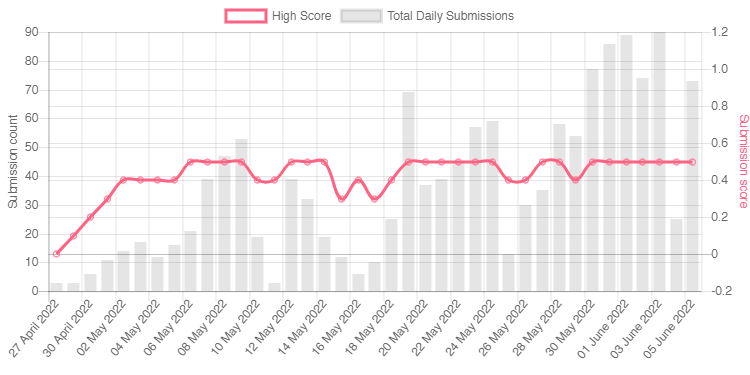}
\caption{Illustration of the trend of number of daily submissions and their scores during the entire phase of the competition.} 
\label{fig:analytics}
\end{figure*}

The competition was active for 52 days from April 27, 2022 through to June 5, 2022. The competition attracted a total of 120 global teams with 1492 submissions. Illustration of the trend of number of daily submissions and their scores during the entire phase of the competition is shown in \autoref{fig:analytics}.
Interestingly, over 93.57\% of submissions were recorded on or after the fourth week, and over 80.63\% of the submissions were recorded during the second half. It can be seen from the graph that during the third week, the number of submissions per day increased gradually to about 40. Since the fourth week, the challenge received an average 45 submissions per day. There was at least one submission in the second half of the challenge with score greater than $0.45$ making their way to the top 10 in the leaderboard with some exceptions.

\begin{figure*}[tb]
\centering
\includegraphics[width=\textwidth]{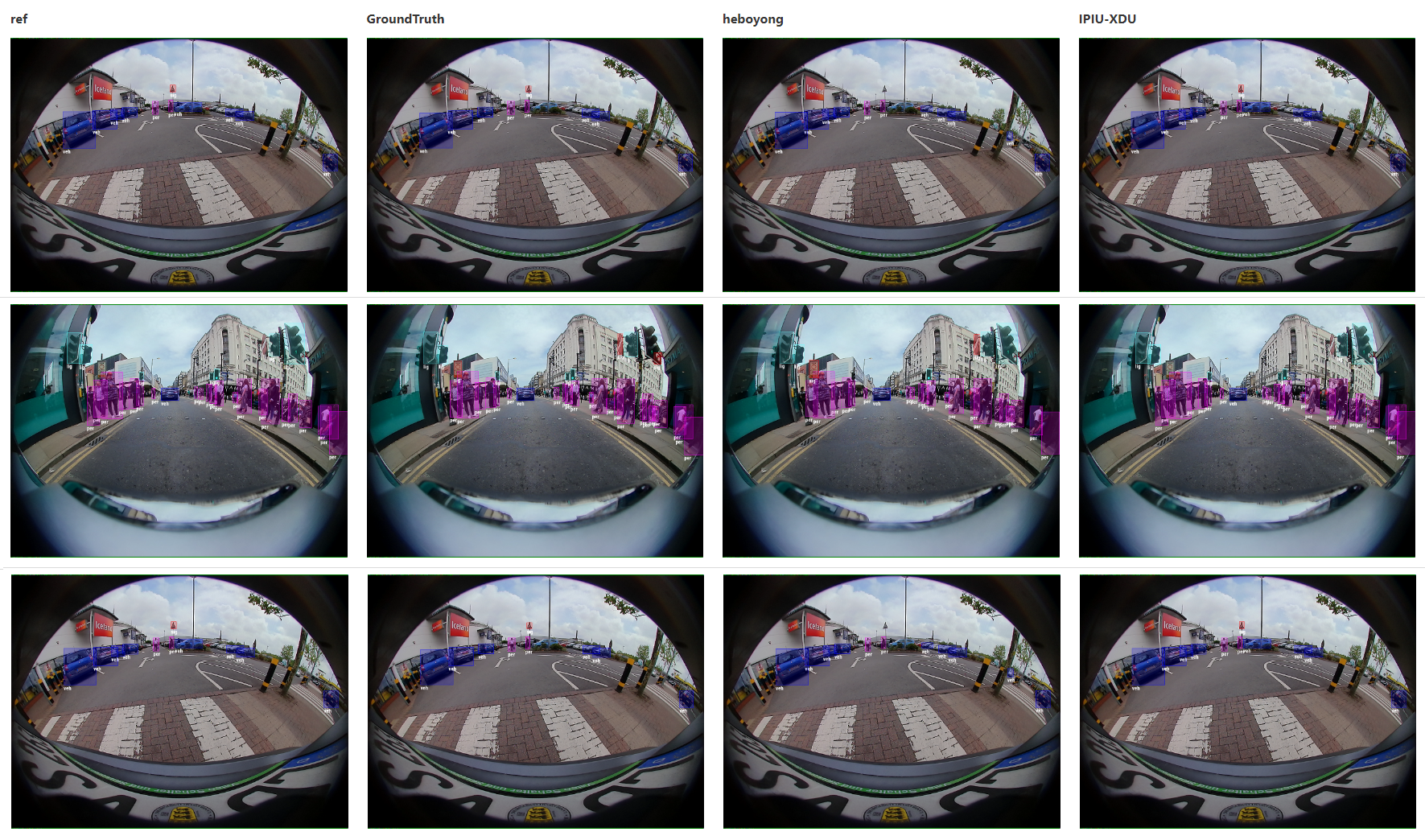}
\caption{Object detection 2D bounding boxes predicted by top 3 teams compared with the reference for 3 randomly picked images. Left to Right: Reference, Team GroundTruth (winner), Team heboyong (second place) and Team IPIU-XDU (third place).} 
\label{fig:pred-comparison}
\end{figure*}

\subsection{Methods}

\subsubsection{Winning Team}
Team \textit{GroundTruth} finished in first place with a score of 0.51 (Vehicles 0.67, Person 0.58, Bicycle 0.44, Traffic Light 0.50, Traffic Sign 0.33) with their Multi-Head Self-Attention (MHSA) Dark Blocks approach. 
Xiaoqiang Lu, Tong Gou, Yuxing Li, Hao Tan, Guojin Cao, and Licheng Jiao
affiliated to 
Guangzhou Institute of Technology, Key Laboratory of Intelligent Perception and Image Understanding of Ministry of Education, Xidian University, and School of Water Conservancy and Hydropower, Xi’an University of Technology belonged to this team. In the individual class scores, they achieved first place for Bicycle class, and second place for the rest of the classes. They adapt the original CSP Darknet \cite{cspdarknet_bochkovskiy2020yolov4} backbone by introducing MHSA layers replacing the CSP bottleneck and with 3x3 and 1x1 spatial convolutional layers. They use a weighted Bidirectional Feature Pyramid Network (BiFPN) \cite{bifpn_tan2020efficientdet} to process multi-scale features. They further improve the detection accuracy using Test Time Augmentation (TTA) and Model Soups \cite{modelsoups_wortsman2022model}. In more detail, multiple augmented copies of each image is generated during inference and bounding boxes predicted by the network for all such copies are returned. They additionally train Scaled-YOLOv4 \cite{scaledyolov4_wang2021scaled} models and use Model Soups ensemble method too enhance the predictions. This resulted in achieving the top score of 0.51.

\begin{table*}
\centering
\begin{tabular}{rlllllll}
\toprule
\# & User              & Score (mAP) $\uparrow$ & Vehicles  $\uparrow$ & Person  $\uparrow$ & Bicycle $\uparrow$ & Traffic Light $\uparrow$ & Traffic Sign $\uparrow$ \\\midrule
{\color{green1} \textbf{1}} & {\color{green1} \textbf{GroundTruth}} & {\color{green1} \textbf{0.51 (1)}} & {\color{green1} \textbf{0.67 (2)}} & {\color{green1} \textbf{0.58 (2)}} & {\color{green1} \textbf{0.44 (1)}} & {\color{green1} \textbf{0.50 (2)}} & {\color{green1} \textbf{0.33 (2)}} \\
{\color{green2} \textbf{2}} & {\color{green2} \textbf{heboyong}}    & {\color{green2} \textbf{0.50 (2)}} & {\color{green2} \textbf{0.67 (1)}} & {\color{green2} \textbf{0.58 (1)}} & {\color{green2} \textbf{0.37 (6)}} & {\color{green2} \textbf{0.53 (1)}} & {\color{green2} \textbf{0.36 (1)}} \\
{\color{green3} \textbf{3}} & {\color{green3} \textbf{IPIU-XDU}}    & {\color{green3} \textbf{0.49 (3)}} & {\color{green3} \textbf{0.67 (3)}} & {\color{green3} \textbf{0.58 (3)}} & {\color{green3} \textbf{0.40 (2)}} & {\color{green3} \textbf{0.50 (3)}} & {\color{green3} \textbf{0.32 (3)}} \\
4                        & miaodq            & 0.49 (4)    & 0.67 (4)  & 0.57 (5)  & 0.40 (3)  & 0.50 (5)      & 0.29 (6)     \\
5                        & xa\_ever          & 0.47 (5)    & 0.66 (6)  & 0.58 (4)  & 0.37 (5)  & 0.49 (6)      & 0.25 (15)    \\
6                        & chenwei           & 0.47 (6)    & 0.66 (9)  & 0.56 (11) & 0.33 (11) & 0.50 (4)      & 0.29 (5)     \\
7                        & BingDwenDwen      & 0.47 (7)    & 0.67 (5)  & 0.57 (7)  & 0.35 (8)  & 0.47 (11)     & 0.28 (11)    \\
8                        & pangzihei         & 0.47 (8)    & 0.65 (10) & 0.57 (6)  & 0.36 (7)  & 0.49 (7)      & 0.26 (14)    \\
9                        & Charles           & 0.46 (9)    & 0.66 (8)  & 0.56 (9)  & 0.31 (12) & 0.49 (9)      & 0.29 (7)     \\
10                       & zx                & 0.46 (10)   & 0.65 (11) & 0.56 (10) & 0.37 (4)  & 0.46 (12)     & 0.27 (12)    \\
11                       & qslb              & 0.46 (11)   & 0.65 (12) & 0.57 (8)  & 0.30 (15) & 0.49 (8)      & 0.28 (10)    \\
12                       & Liyuanhao         & 0.46 (12)   & 0.66 (7)  & 0.56 (12) & 0.29 (16) & 0.48 (10)     & 0.28 (9)     \\
13                       & keeply            & 0.44 (13)   & 0.62 (16) & 0.52 (13) & 0.34 (10) & 0.42 (15)     & 0.29 (8)     \\
14                       & icecreamztq       & 0.43 (14)   & 0.63 (13) & 0.51 (16) & 0.28 (17) & 0.44 (13)     & 0.26 (13)    \\
15                       & asdada            & 0.42 (15)   & 0.61 (19) & 0.52 (14) & 0.35 (9)  & 0.41 (17)     & 0.23 (17)    \\
16                       & determined\_dhaze & 0.42 (16)   & 0.62 (17) & 0.49 (19) & 0.30 (14) & 0.42 (14)     & 0.29 (4)     \\
17                       & zhuzhu            & 0.42 (17)   & 0.61 (20) & 0.52 (14) & 0.35 (9)  & 0.41 (17)     & 0.23 (17)    \\
18                       & msc\_1            & 0.42 (18)   & 0.63 (15) & 0.51 (17) & 0.30 (13) & 0.41 (18)     & 0.24 (16)    \\
19                       & hgfwgwf           & 0.41 (19)   & 0.61 (22) & 0.52 (15) & 0.35 (9)  & 0.37 (21)     & 0.21 (19)    \\
20                       & cscbs             & 0.41 (20)   & 0.61 (21) & 0.46 (23) & 0.35 (9)  & 0.41 (17)     & 0.23 (17)    \\\bottomrule
\end{tabular}
\caption{Snapshot of the challenge leaderboard illustrating the top twenty participants based on the \textit{mAP Score} metric. Top three participants are highlighted in shades of green.  }
\label{tab:leaderboard}
\end{table*}

\subsubsection{Second Place}
Team \textit{heboyong} finished in second place with a score of 0.50 (Vehicles 0.67, Person 0.58, Bicycle 0.37, Traffic Light 0.53, Traffic Sign 0.36) using Swin Transformers.
He Boyong, Guo Weijie, Ye Qianwen, and Li Xianjiang
affiliated to 
Xiamen University belonged to this team. In the individual class scores, they achieved first place for all classes except Bicycle class for which they ranked sixth. They use Cascade RCNN \cite{cascadercnn_cai2018cascade} with a SwinTransformer \cite{swinv1_liu2021swin} backbone. They utilize CBNetv2 \cite{liang2021cbnetv2} network architecture to improve accuracy of the Swin Transformer backbone without retraining and use Seesaw Loss \cite{wang2021seesaw} to address the long-tailed problem that occurs between the categories of quantitative imbalance. They use MixUp \cite{zhang2018mixup} and Albumentation \cite{buslaev2020albumentations} along with random scaling and random crop for data augmentation. They additionally train models using the circular cosine learning rate setting for an additional 12 epochs and average all models using the Stochastic Weight Averaging \cite{swa_izmailov2018averaging} method to acquire a final model that is more robust and accurate. Finally, in the inference stage, they use Soft-NMS \cite{softnms_bodla2017soft}, multi-scale augmentation, and flip augmentation to further enhance the results.

\subsubsection{Third Place}
Team \textit{IPIU-XDU} finished in third place with a score of 0.49 (Vehicles 0.67, Person 0.58, Bicycle 0.40, Traffic Light 0.50, Traffic Sign 0.32) using Swin Transformer networks. 
Chenghui Li, Chao Li, Xiao Tan, Zhongjian Huang, and Yuting Yang
affiliated to
Hangzhou Institute of Technology, Xidian University belonged to this team. In the individual class scores, they achieved second place for Bicycle class, and third place for the rest of the classes. They utilize Swin Transformer v2 \cite{swinv2_liu2022swin} with HTC++ (Hybrid Task Cascade) \cite{htc_chen2019hybrid} \cite{swinv1_liu2021swin} to detect objects in fisheye images. They use multi-scale training with ImageNet-22K \cite{deng2009_imagenet} pretrained backbone with its learning rate set to one-tenth of that of the head and use Soft-NMS \cite{softnms_bodla2017soft} for inference. An ensemble architecture is used to boost the scores, the confidence scores of bounding boxes proposed by each model is used and averaged using Weighted Boxes Fusion \cite{wbf_solovyev2021weighted}.

\subsection{Results and Discussion}

Team \textit{GroundTruth}, with a lead score of 0.51 (Vehicles 0.67, Person 0.58, Bicycle 0.44, Traffic Light 0.50, Traffic Sign 0.33), was announced as the winner on 6th June 2022.
In Table \autoref{tab:leaderboard}, we showcase the challenge leaderboard with details of the top twenty team participants. 
The winning team \textit{GroundTruth} presented their method virtually in the OmniCV Workshop, CVPR 2022, held on June 20, 2022. In \autoref{fig:pred-comparison}, we illustrate outputs of the top 3 teams from randomly picked samples. 

\section{Conclusion} \label{sec:conc}
In this paper, we discussed the results of the fisheye object detection challenge hosted at our CVPR OmniCV workshop 2022. Spatially variant radial distortion makes the object detection task quite challenging. In addition, bounding boxes are sub-optimal representations of objects particularly at periphery which have a curved box shape. Most solutions submitted did not explicitly made use of the radial distortion model to exploit the known camera model. The top performing methods made use of transformers which seem to learn the radial distortion implicitly. Extensive augmentation methods were also used. 
We have started accepting submissions again keeping the challenge open to everyone to encourage further research and novel solutions to fisheye object detection. In our future work, we plan to organize similar workshop challenges on fisheye camera multi-task learning.

\section*{Acknowledgments}

Woodscape OmniCV 2022 Challenge
was supported
in part
by \href{https://www.sfi.ie}{Science Foundation Ireland} grant 13/RC/2094 to \href{https://www.lero.ie}{Lero - the Irish Software Research Centre} and grant 16/RI/3399.

\bibliographystyle{IEEEtran}
\bibliography{IEEEabrv, references}
\end{document}